\pdfoutput=1

\documentclass[11pt]{article}

\usepackage[preprint]{acl}

\usepackage{times}
\usepackage{latexsym}
\usepackage[T1]{fontenc}
\usepackage[utf8]{inputenc}
\usepackage{microtype}
\usepackage{inconsolata}
\usepackage{graphicx}
\usepackage{booktabs}
\usepackage{multirow}
\usepackage{amsmath}
\usepackage{amssymb}
\usepackage{amsthm}
\usepackage{bbding}
\usepackage{xspace}

\usepackage{tikz}
\usetikzlibrary{calc, arrows, arrows.meta, intersections, patterns, positioning, shapes, fadings, through, decorations.pathreplacing}

\newcommand{\fwname}{\textsc{LlamaFactory}\xspace}
\newcommand{\ifname}{\textsc{LlamaBoard}\xspace}
\newcommand{\hfname}{Hugging Face\xspace}
\newcommand{\yes}{\Checkmark}
\newcommand{\noh}{\XSolidBrush}
\newcommand{\eg}{\textit{e}.\textit{g}.}

\title{\fwname: Unified Efficient Fine-Tuning of 100+ Language Models}
\author{
Yaowei Zheng\textsuperscript{1}, Richong Zhang\textsuperscript{1}\thanks{Corresponding author}, Junhao Zhang\textsuperscript{1}, Yanhan Ye\textsuperscript{1}, \\
{\bf Zheyan Luo\textsuperscript{1}, Zhangchi Feng\textsuperscript{1}, Yongqiang Ma\textsuperscript{2}} \\
\textsuperscript{1}School of Computer Science and Engineering, Beihang University, China \\
\textsuperscript{2}School of Software and Microelectronics, Peking University, China \\
{\tt\small \{hiyouga,zhang.jh,yeyanhan,akamya,zcmuller\}@buaa.edu.cn},  {\tt\small zhangrc@act.buaa.edu.cn}, {\tt\small codingma@pku.edu.cn} \\
Demonstration video: {\bf\url{https://youtu.be/W29FgeZEpus}}
}

\begin{document}

\maketitle

\begin{abstract}

Efficient fine-tuning is vital for adapting large language models (LLMs) to downstream tasks. However, it requires non-trivial efforts to implement these methods on different models. We present \fwname, a unified framework that integrates a suite of cutting-edge efficient training methods. It provides a solution for flexibly customizing the fine-tuning of 100+ LLMs without the need for coding through the built-in web UI \ifname. We empirically validate the efficiency and effectiveness of our framework on language modeling and text generation tasks. It has been released at \url{https://github.com/hiyouga/LLaMA-Factory} and received over 25,000 stars and 3,000 forks.

\end{abstract}

\section{Introduction}

Large language models (LLMs) \citep{zhao2023survey} present remarkable reasoning capabilities and empower a wide range of applications, such as question answering \citep{jiang2023reasoninglm}, machine translation \citep{wang2023document,jiao2023parrot}, and information extraction \citep{jiao2023instruct}. Subsequently, a substantial number of LLMs are developed and accessible through open-source communities. For example, \hfname's open LLM leaderboard \citep{beeching2023leaderboard} boasts over 5,000 models, offering convenience for individuals seeking to leverage the power of LLMs.

Fine-tuning extremely large number of parameters with limited resources becomes the main challenge of adapting LLM to downstream tasks. A popular solution is efficient fine-tuning \citep{houlsby2019adapter,hu2022lora,dettmers2023qlora}, which reduces the training cost of LLMs when adapting to various tasks. However, the community contributes various methods for efficient fine-tuning, lacking a systematic framework that adapts and unifies these methods to different LLMs and provides a friendly interface for user customization.

To address the above problems, we develop \fwname, a framework that democratizes the fine-tuning of LLMs. It unifies a variety of efficient fine-tuning methods through scalable modules, enabling the fine-tuning of hundreds of LLMs with minimal resources and high throughput. In addition, it streamlines commonly used training approaches, including generative pre-training \citep{radford2018gpt}, supervised fine-tuning (SFT) \citep{wei2022instruction}, reinforcement learning from human feedback (RLHF) \cite{ouyang2022instructgpt}, and direct preference optimization (DPO) \cite{rafailov2023dpo}. Users can leverage command-line or web interfaces to customize and fine-tune their LLMs with minimal or no coding effort.

\fwname consists of three main modules: {\bf\em Model Loader}, {\bf\em Data Worker} and {\bf\em Trainer}. We minimize the dependencies of these modules on specific models and datasets, allowing the framework to flexibly scale to hundreds of models and datasets. Concretely, we first establish a model registry where the {\em Model Loader} can precisely attach adapters to the pre-trained models by identifying exact layers. Then we develop a data description specification that allows the {\em Data Worker} to gather datasets by aligning corresponding columns. Furthermore, we provide plug-and-play implementations of state-of-the-art efficient fine-tuning methods that enable the {\em Trainer} to activate by replacing default ones. Our design allows these modules to be reused across different training approaches, significantly reducing the integration costs.

\fwname is implemented with PyTorch \citep{paszke2019pytorch} and significantly benefits from open-source libraries, such as Transformers \citep{wolf2020transformers}, PEFT \citep{mangrulkar2022peft}, and TRL \citep{vonwerra2022trl}. On the basis, we provide an out-of-the-box framework with a higher level of abstraction. Additionally, we build \ifname with Gradio \citep{abid2019gradio}, enabling fine-tuning LLMs with no coding efforts required.

\fwname is open-sourced under the Apache-2.0 license. It has already garnered over 25,000 stars and 3,000 forks on the GitHub, and hundreds of open-source models have been built upon \fwname on the \hfname Hub\footnote{\url{https://huggingface.co/models?other=llama-factory}}. For example, \citet{truong2024crossing} build GemSUra-7B based on \fwname, revealing the cross-lingual abilities of Gemma \citep{team2024gemma}. Furthermore, dozens of studies have utilized our framework to explore LLMs \citep{wang2023esrl,yu2023textcls,bhardwaj2024language}.

\section{Related Work}

\begin{table}[t]
\centering
\resizebox{0.99\columnwidth}{!}{%
\begin{tabular}{cccccc}
\toprule
& \fwname & FastChat & LitGPT & LMFlow & Open-Instruct \\
\midrule
LoRA   & \yes & \yes & \yes & \yes & \yes \\
QLoRA  & \yes & \yes & \yes & \yes & \yes \\
DoRA   & \yes & & & & \\
LoRA+  & \yes & & & & \\
PiSSA & \yes & & & & \\
GaLore & \yes & \yes & & \yes & \yes \\
BAdam & \yes & & & & \\
\midrule
Flash attention & \yes & \yes & \yes & \yes & \yes \\
S$^2$ attention & \yes & & & & \\
Unsloth & \yes & & \yes & & \\
DeepSpeed & \yes & \yes & \yes & \yes & \yes \\
\midrule
SFT & \yes & \yes & \yes & \yes & \yes \\
RLHF & \yes & & & \yes & \\
DPO & \yes & & & & \yes \\
KTO & \yes & & & & \\
ORPO & \yes & & & & \\
\bottomrule
\end{tabular}%
}
\caption{Comparison of features in \fwname with popular frameworks of fine-tuning LLMs.}
\label{tab:checklist}
\end{table}

With the rapid increase in demand for fine-tuning LLMs, numerous frameworks for adapting LLMs to specific purposes have been developed. LLaMA-Adapter \citep{zhang2024llamaadapter} efficiently fine-tunes the Llama model \citep{touvron2023llama} using a zero-initialized attention. FastChat \citep{zheng2023fastchat} is a framework focused on training and evaluating LLMs for chat completion purposes. LitGPT \citep{lighting2023litgpt} provides the implementation of generative models and supports various training methods. Open-Instruct \citep{wang2023openinstruct} provides recipes for training instruct models. Colossal AI \citep{li2023colossalai} takes advanced parallelism strategies for distributed training. LMFlow \citep{diao2024lmflow} supports training LLMs for specialized domains or tasks. GPT4All \citep{anand2023gpt4all} allows LLMs to run on consumer devices, while also providing fine-tuning capabilities. Compared with existing competitive frameworks, \fwname supports a broader range of efficient fine-tuning techniques and training approaches. We list the features among representative frameworks in Table~\ref{tab:checklist}.

\section{Efficient Fine-Tuning Techniques}

Efficient LLM fine-tuning techniques can be divided into two main categories: those focused on optimization and those aimed at computation. The primary objective of efficient optimization techniques is to fine-tune the parameters of LLMs while keeping costs to a minimum. On the other hand, efficient computation methods seek to decrease the time or space for the required computation in LLMs. The methods included in \fwname are listed in Table~\ref{tab:methods}. We will present these efficient fine-tuning techniques and show the substantial efficiency improvement achieved by incorporating them into our framework in the following sections.

\begin{table}[t]
\centering
\resizebox{0.99\columnwidth}{!}{%
\begin{tabular}{lccccccc}
\toprule
 & Freeze-tuning & GaLore & LoRA & DoRA & LoRA+ & PiSSA \\
\midrule
Mixed precision & \yes & \yes & \yes & \yes & \yes & \yes \\
Checkpointing & \yes & \yes & \yes & \yes & \yes & \yes \\
Flash attention & \yes & \yes & \yes & \yes & \yes & \yes \\
S$^2$ attention & \yes & \yes & \yes & \yes & \yes & \yes \\
Quantization & \noh & \noh & \yes & \yes & \yes & \yes \\
Unsloth & \noh & \noh & \yes & \yes & \yes & \yes \\
\bottomrule
\end{tabular}%
}
\caption{Compatibility between the fine-tuning techniques featured in \fwname.}
\label{tab:methods}
\end{table}

\subsection{Efficient Optimization}

Firstly, we provide an overview of the efficient optimization techniques utilized in \fwname. The freeze-tuning method \citep{houlsby2019adapter} involves freezing a majority of parameters while fine-tuning the remaining parameters in a small subset of decoder layers. Another method called gradient low-rank projection (GaLore) \citep{zhao2024galore} projects gradients into a lower-dimensional space, facilitating full-parameter learning in a memory-efficient manner. Similarly, BAdam \citep{luo2024badam} leverages block coordinate descent (BCD) to efficiently optimize the extensive parameters. On the contrary, the low-rank adaptation (LoRA) \citep{hu2022lora} method freezes all pre-trained weights and introduces a pair of trainable low-rank matrices to the designated layer. When combined with quantization, this approach is referred to as QLoRA \citep{dettmers2023qlora}, which additionally reduces the memory usage. DoRA \citep{liu2024dora} breaks down pre-trained weights into magnitude and direction components and updates directional components for enhanced performance. LoRA+ \citep{hayou2024loraplus} is proposed to overcome the sub-optimality of LoRA. PiSSA \citep{meng2024pissa} initializes adapters with the principal components of the pre-trained weights for faster convergence.

\subsection{Efficient Computation}

In \fwname, we integrate a range of techniques for efficient computation. Commonly utilized techniques encompass mixed precision training \citep{micikevicius2018mixed} and activation checkpointing \citep{chen2016checkpointing}. Drawing insights from the examination of the input-output (IO) expenses of the attention layer, flash attention \citep{dao2022flashattention} introduces a hardware-friendly approach to enhance attention computation. S$^2$ attention \citep{chen2024longlora} tackles the challenge of extended context with shifted sparse attention, thereby diminishing memory usage in fine-tuning long-context LLMs. Various quantization strategies \citep{dettmers2022int8,frantar2023gptq,lin2023awq,egiazarian2024aqlm} decrease memory requirements in large language models (LLMs) by utilizing lower-precision representations for weights. Nevertheless, the fine-tuning of quantized models is restricted to the adapter-based techniques like LoRA \citep{hu2022lora}. Unsloth \citep{han2023unsloth} incorporates Triton \citep{tillet2019triton} for implementing the backward propagation of LoRA, which reduces floating-point operations (FLOPs) during gradient descent and leads to expedited LoRA training.

\fwname seamlessly combines these techniques into a cohesive structure to enhance the efficiency of LLM fine-tuning. This results in a reduction of the memory footprint from 18 bytes per parameter during mixed precision training \citep{micikevicius2018mixed} or 8 bytes per parameter in half precision training \citep{le2022bloom} to only 0.6 bytes per parameter. Further elaboration on the components in \fwname will be provided in the subsequent section.

\section{\fwname Framework}

\fwname consists of three main modules: {\em Model Loader}, {\em Data Worker}, and {\em Trainer}. The {\em Model Loader} manipulates various model architectures for fine-tuning, supporting both large language models (LLMs) and vision language models (VLMs). The {\em Data Worker} processes data from different tasks through a well-designed pipeline, supporting both single-turn and multi-turn dialogues. The {\em Trainer} applies efficient fine-tuning techniques to different training approaches, supporting pre-training, instruction tuning and preference optimization. Beyond that, \ifname provides a friendly visual interface to access these modules, enabling users to configure and launch individual LLM fine-tuning instance codelessly and monitor the training status synchronously. We illustrate the relationships between these modules and the overall architecture of \fwname in Figure~\ref{fig:arch}.

\begin{figure}
\centering
\resizebox{.99\columnwidth}{!}{\begin{tikzpicture}


\tikzset{  
  every path/.style={line width=1pt},  
}

\tikzstyle{title}   = [midway, text centered, text=white, font=\Large]
\tikzstyle{content} = [midway, text centered]

\tikzstyle{textBox} = [draw=none, fill opacity=1]

\tikzstyle{yellowBox} = [draw=black!50!yellow, rounded corners, fill=black!10!yellow!10]
\tikzstyle{blueBox}   = [draw=black!60!blue,   rounded corners, fill=black!10!blue!5]
\tikzstyle{greenBox}  = [draw=black!60!green,  rounded corners, fill=black!10!green!5]
\tikzstyle{orangeBox} = [draw=black!60!orange, rounded corners, fill=black!10!orange!5]

\tikzstyle{yellowTitle} = [draw=none, rounded corners, fill=black!50!yellow]
\tikzstyle{blueTitle}   = [draw=none, rounded corners, fill=black!60!blue]
\tikzstyle{greenTitle}  = [draw=none, rounded corners, fill=black!60!green]
\tikzstyle{orangeTitle} = [draw=none, rounded corners, fill=black!60!orange]

\tikzstyle{yellowNode} = [draw=black!50!yellow, rounded corners, fill=black!10!yellow!10]
\tikzstyle{greenNode}  = [draw=black!60!green,  rounded corners, fill=black!10!green!10]
\tikzstyle{orangeNode} = [draw=black!60!orange, rounded corners, fill=black!10!orange!10]
\tikzstyle{blueDiv}    = [draw=black!60!blue,   rounded corners, fill=black!20!blue!15]
\tikzstyle{blueNode}   = [draw=black!60!blue,   rounded corners, fill=black!10!blue!10]

\draw[yellowBox] (0,9.8) rectangle +(10,1.8);
\draw[yellowTitle] (3.25,11.25) rectangle +(3.5,0.8) node[title] {\texttt{\textbf{LlamaBoard}}};
\draw[yellowNode] (0.2,10.15) rectangle +(4.6,0.8) node[content] {\textsf{Experiment Configurator}};
\draw[yellowNode] (5.2,10.15) rectangle +(4.6,0.8) node[content] {\textsf{Training Status Monitor}};

\draw[blueBox] (0,5.1) rectangle +(10,3.8);
\draw[blueTitle] (3.25,8.5) rectangle +(3.5,0.8) node[title] {\texttt{\textbf{Trainer}}};
\draw[blueDiv] (0.2,5.45) rectangle +(4.6,2.8);
\draw[blueDiv] (5.2,5.45) rectangle +(4.6,2.8);
\draw[textBox] (0.2,7.45) rectangle +(4.6,0.8) node[content] {\textsf{\textbf{Optimization}}};
\draw[textBox] (5.2,7.45) rectangle +(4.6,0.8) node[content] {\textsf{\textbf{Approaches}}};

\draw[blueNode] (0.45,6.65) rectangle +(1.9,0.8) node[content] {\textsf{LoRA}};
\draw[blueNode] (2.65,6.65) rectangle +(1.9,0.8) node[content] {\textsf{PiSSA}};
\draw[blueNode] (0.45,5.65) rectangle +(1.9,0.8) node[content] {\textsf{GaLore}};
\draw[blueNode] (2.65,5.65) rectangle +(1.9,0.8) node[content] {\textsf{BAdam}};

\draw[blueNode] (5.45,6.65) rectangle +(1.9,0.8) node[content] {\textsf{Pre-train}};
\draw[blueNode] (7.65,6.65) rectangle +(1.9,0.8) node[content] {\textsf{SFT}};
\draw[blueNode] (5.45,5.65) rectangle +(1.9,0.8) node[content] {\textsf{RLHF}};
\draw[blueNode] (7.65,5.65) rectangle +(1.9,0.8) node[content] {\textsf{DPO}};

\draw[greenBox]  (0,1.3)   rectangle +(4.8,2.9);
\draw[orangeBox] (5.2,1.3) rectangle +(4.8,2.9);

\draw[greenTitle]  (0.65,3.8) rectangle +(3.5,0.8) node[title] {\texttt{\textbf{Model Loader}}};
\draw[orangeTitle] (5.85,3.8) rectangle +(3.5,0.8) node[title] {\texttt{\textbf{Data Worker}}};

\draw[greenNode] (0.2,2.65) rectangle +(2.4,0.8) node[content] {\textsf{Initialization}};
\draw[greenNode] (2.8,2.65)   rectangle +(1.8,0.8) node[content] {\textsf{Patches}};
\draw[greenNode] (0.2,1.6)  rectangle +(2.4,0.8) node[content] {\textsf{Quantization}};
\draw[greenNode] (2.8,1.6)    rectangle +(1.8,0.8) node[content] {\textsf{Adapters}};

\draw[orangeNode] (7.4,2.65) rectangle +(2.4,0.8) node[content] {\textsf{Aligning}};
\draw[orangeNode] (5.4,2.65) rectangle +(1.8,0.8) node[content] {\textsf{Loading}};
\draw[orangeNode] (7.4,1.6)  rectangle +(2.4,0.8) node[content] {\textsf{Preprocess}};
\draw[orangeNode] (5.4,1.6)  rectangle +(1.8,0.8) node[content] {\textsf{Merging}};

\draw[greenBox]  (0,-0.2)   rectangle +(4.8,0.8) node[content] {\textsf{Pre-Trained Models}};
\draw[orangeBox] (5.2,-0.2) rectangle +(4.8,0.8) node[content] {\textsf{Conversational Datasets}};

\tikzstyle{nodeline} = [draw=black, arrows={-Latex[scale length=0.45, scale width=0.7]}, line width=2.25pt]

\draw[nodeline] (2.5,10.15) -- +(0,-1.25);
\draw[nodeline] (7.5,8.9) -- +(0,1.25);

\draw[nodeline] (2.4,4.6) -- +(0,0.85);
\draw[nodeline] (7.6,4.6) -- +(0,0.85);

\draw[nodeline] (2.4,0.6) -- +(0,0.7);
\draw[nodeline] (7.6,0.6) -- +(0,0.7);

\end{tikzpicture}}
\caption{The architecture of \fwname.}
\label{fig:arch}
\end{figure}

\subsection{Model Loader}

This section initially presents the four components in {\em Model Loader}: model initialization, model patching, model quantization, and adapter attaching, followed by a description of our approach of adapting to a wide range of devices by handling the parameter floating-point precision during fine-tuning.

\paragraph{Model Initialization}

We utilize the {\em Auto Classes} of Transformers \citep{wolf2020transformers} to load pre-trained models and initialize parameters. Specifically, we load the vision language models using the {\tt AutoModelForVision2Seq} class while the rest are loaded using the {\tt AutoModelForCausalLM} class. The tokenizer is loaded using the {\tt AutoTokenizer} class along with the model. In cases where the vocabulary size of the tokenizer exceeds the capacity of the embedding layer, we resize the layer and initialize new parameters with noisy mean initialization. To determine the scaling factor for RoPE scaling \citep{chen2023extending}, we compute it as the ratio of the maximum input sequence length to the context length of the model.

\paragraph{Model Patching}

To enable the S$^2$ attention, we employ a monkey patch to replace the forward computation of models. However, we use the native class to enable flash attention as it has been widely supported since Transformers 4.34.0. To prevent excessive partitioning of the dynamic layers, we set the mixture-of-experts (MoE) blocks as leaf modules when we optimize the MoE models under DeepSpeed ZeRO stage-3 \citep{rasley2020deepspeed}.

\paragraph{Model Quantization}

Dynamically quantizing models to 8 bits or 4 bits with LLM.int8 \citep{dettmers2022int8} can be performed through the bitsandbytes library \citep{dettmers2021bitsandbytes}. For 4-bit quantization, we utilize the double quantization and 4-bit normal float as QLoRA \citep{dettmers2023qlora}. We also support fine-tuning the models quantized by the post-training quantization (PTQ) methods, including GPTQ \citep{frantar2023gptq}, AWQ \citep{lin2023awq}, and AQLM \citep{egiazarian2024aqlm}. Note that we cannot directly fine-tune the quantized weights; thus, the quantized models are only compatible with adapter-based methods.

\paragraph{Adapter Attaching}

We automatically identify the appropriate layers to attach adapters through traversing the model layers. The low-rank adapters are attached to all the linear layers for a better convergence as suggested by \citep{dettmers2023qlora}. The PEFT \citep{mangrulkar2022peft} library provides an extremely convenient way to implement the adapter-based methods such as LoRA \citep{hu2022lora}, rsLoRA \citep{kalajdzievski2023rslora}, DoRA \citep{liu2024dora} and PiSSA \citep{meng2024pissa}. We replace the backward computation with the one of Unsloth \citep{han2023unsloth} to accelerate the training. To perform reinforcement learning from human feedback (RLHF), a value head layer is appended on the top of the transformer model, mapping the representation of each token to a scalar.

\paragraph{Precision Adaptation}

We handle the floating-point precision of pre-trained models based on the capabilities of computing devices. For NVIDIA GPUs, we adopt bfloat16 precision if the computation capability is 8.0 or higher. Otherwise, float16 is adopted. Besides, we adopt float16 for Ascend NPUs and AMD GPUs and float32 for non-CUDA devices. In mixed precision training, we set all trainable parameters to float32 for training stability. Nevertheless, we retain the trainable parameters as bfloat16 in half precision training.

\subsection{Data Worker}

We develop a data processing pipeline, including dataset loading, dataset aligning, dataset merging and dataset pre-processing. It standardizes datasets of different tasks into a unified format, enabling us to fine-tune models on datasets in various formats.

\paragraph{Dataset Loading}

We utilize the Datasets \citep{lhoest2021datasets} library to load the data, which allows the users to load remote datasets from the \hfname Hub or read local datasets via scripts or through files. The Datasets library significantly reduces memory overhead during data processing and accelerates sample querying using Arrow \citep{apache2016arrow}. By default, the whole dataset is downloaded to local disk. However, if a dataset is too large to be stored, our framework provides dataset streaming to iterate over it without downloading.

\begin{table}[t]
\centering
\resizebox{0.99\columnwidth}{!}{%
\begin{tabular}{lp{7.5cm}}
\toprule
Plain text & [\{"text": "..."\}, \{"text": "..."\}] \\
Alpaca-like data & [\{"instruction": "...", "input": "...", "output": "..."\}] \\
ShareGPT-like data & [\{"conversations": [\{"from": "human", "value": "..."\}, \{"from": "gpt", "value": "..."\}]\}]\\
Preference data & [\{"instruction": "...", "input": "...", "output": ["...", "..."]\}] \\
\midrule
Standardized data & \{"prompt": [\{"role": "...", "content": "..."\}], \\
& "response": [\{"role": "...", "content": "..."\}], \\
& "system": "...", "tools": "...", "images": ["..."]\} \\
\bottomrule
\end{tabular}%
}
\caption{Dataset structures in \fwname.}
\label{tab:format}
\end{table}

\paragraph{Dataset Aligning}

To unify the dataset format, we design a data description specification to characterize the structure of datasets. For example, the alpaca dataset has three columns: instruction, input and output \citep{taori2023alpaca}. We convert the dataset into a standard structure that is compatible with various tasks according to the data description specification. Some examples of dataset structures are shown in Table~\ref{tab:format}.

\paragraph{Dataset Merging}

The unified dataset structure provides an efficient approach for merging multiple datasets. For the datasets in non-streaming mode, we simply concatenate them before the datasets are shuffled during training. However, in streaming mode,  simply concatenating the datasets impedes data shuffling. Therefore, we offer methods to alternately read the data from different datasets.

\paragraph{Dataset Pre-processing}

\fwname is designed for fine-tuning the text generative models, which is primarily used in chat completion. Chat template is a crucial component in these models, because it is highly related to the instruction-following abilities of these models. Therefore, we provide dozens of chat templates that can be automatically chosen according to the model type. We encode the sentence after applying the chat template using the tokenizer. By default, we only compute loss on the completions, while the prompts are disregarded \citep{taori2023alpaca}. Optionally, we can utilize sequence packing \citep{krell2021packing} to reduce the training time, which is automatically enabled when performing generative pre-training.

\subsection{Trainer}

\paragraph{Efficient Training}

We integrate state-of-the-art efficient fine-tuning methods, including LoRA+ \citep{hayou2024loraplus}, GaLore \citep{zhao2024galore} and BAdam \citep{luo2024badam} to the {\em Trainer} by replacing the default components. These fine-tuning methods are independent of the {\em Trainer}, making them easily applicable to various tasks. We utilize the trainers of Transformers \citep{wolf2020transformers} for pre-training and SFT, while adopting the trainers of TRL \citep{vonwerra2022trl} for RLHF and DPO. We also include trainers of the advanced preference optimization methods such as KTO \citep{ethayarajh2024kto} and ORPO \citep{hong2024orpo} from the TRL library. The tailored data collators are leveraged to differentiate trainers of various training approaches. To match the input format of the trainers for preference data, we build $2n$ samples in a batch where the first $n$ samples are chosen examples and the last $n$ samples are rejected examples.

\paragraph{Model-Sharing RLHF}

Allowing RLHF training on consumer devices is crucial for democratizing LLM fine-tuning. However, it is difficult because RLHF training requires four different models. To address this problem, we propose model-sharing RLHF, enabling entire RLHF training with no more than one pre-trained model. Concretely, we first train an adapter and a value head with the objective function for reward modeling, allowing the model to compute reward scores. Then we initialize another adapter and value head and train them with the PPO algorithm \citep{ouyang2022instructgpt}. The adapters and value heads are dynamically switched through the {\tt set\_adapter} and {\tt disable\_adapter} methods of PEFT \citep{mangrulkar2022peft} during training, allowing a single pre-trained model to serve as policy model, value model, reference model, and reward model simultaneously. To the best of our knowledge, this is the first method that supports RLHF training on consumer devices.

\paragraph{Distributed Training}

We can combine the above trainers with DeepSpeed \citep{rasley2020deepspeed,ren2021offload} for distributed training. We adopt data parallelism to fully exploit the ability of computing devices. Leveraging the DeepSpeed ZeRO optimizer, the memory consumption can be further reduced via partitioning or offloading.

\subsection{Utilities}

\paragraph{Model Inference}

During inference time, we reuse the chat template from the {\em Data Worker} to build the model inputs. We offer support for sampling the model outputs using Transformers \citep{wolf2020transformers} and vLLM \citep{kwon2023vllm}, both of which support stream decoding. Additionally, we implement an OpenAI-style API that utilizes the asynchronous LLM engine and paged attention of vLLM, to provide high-throughput concurrent inference services, facilitating the deployment of fine-tuned LLMs into various applications.

\paragraph{Model Evaluation}

We include several metrics for evaluating LLMs, including multiple-choice tasks such as MMLU \citep{hendrycks2021mmlu}, CMMLU \citep{li2023cmmlu}, and C-Eval \citep{huang2023ceval}, as well as calculating text similarity scores like BLEU-4 \citep{papineni2002bleu} and ROUGE \citep{lin2004rouge}. This feature facilitates users to measure the abilities of the fine-tuned models.

\subsection{\ifname: A Unified Interface for \fwname}

\begin{table*}[t]
\centering
\resizebox{1.99\columnwidth}{!}{%
\begin{tabular}{l|rrrr|rrrr|rrrr}
\toprule

&
\multicolumn{4}{|c|}{Gemma-2B} &
\multicolumn{4}{|c|}{Llama2-7B} &
\multicolumn{4}{|c }{Llama2-13B} \\

\midrule

Method &
Trainable & Memory & Throughput & PPL &
Trainable & Memory & Throughput & PPL &
Trainable & Memory & Throughput & PPL \\

&
Params & (GB) & (Tokens/s) & &
Params & (GB) & (Tokens/s) & &
Params & (GB) & (Tokens/s) & \\

\midrule

Baseline &
/ & / & / & 11.83 &
/ & / & / & 7.53 &
/ & / & / & 6.66 \\

Full-tuning &
2.51B & 17.06 & 3090.42 & 10.34 &
6.74B & 38.72 & 1334.72 & 5.56 &
/ & / & / & / \\

Freeze-tuning &
0.33B & 8.10 & 5608.49 & 11.33 &
0.61B & 15.69 & 2904.98 & 6.59 &
0.95B & 29.02 & 1841.46 & 6.56 \\

GaLore &
2.51B & 10.16 & 2483.05 & 10.38 &
6.74B & 15.43 & 1583.77 & 5.88 &
13.02B & 28.91 & 956.39 & \textbf{5.72} \\

LoRA &
0.16B & 7.91 & \textbf{3521.05} & \textbf{10.19} &
0.32B & 16.32 & \textbf{1954.07} & \textbf{5.81} &
0.50B & 30.09 & \textbf{1468.19} & 5.75 \\

QLoRA &
0.16B & \textbf{5.21} & 3158.59 & 10.46 &
0.32B & \textbf{7.52} & 1579.16 & 5.91 &
0.50B & \textbf{12.61} & 973.53 & 5.81 \\

\bottomrule
\end{tabular}%
}
\caption{Comparison of the training efficiency using different fine-tuning methods in \fwname. The best result among GaLore, LoRA and QLoRA of each model is in {\bf bold}.}
\label{tab:efficiency}
\end{table*}

\ifname is a unified user interface based on Gradio \citep{abid2019gradio} that allows users to customize the fine-tuning of LLMs without writing any code. It offers a streamlined model fine-tuning and inference service, enabling users to easily explore the potential of LLMs in their environments. \ifname has the following notable features.

\paragraph{Easy Configuration}

\ifname allows us to customize the fine-tuning arguments through interaction with the web interface. We provide default values for a majority of arguments that are recommended for most users, simplifying the configuration process. Moreover, users can preview the datasets on the web UI to validate them.

\paragraph{Monitorable Training}

During the training process, the training logs and loss curves are visualized and updated in real time, allowing users to monitor the training progress. This feature provides valuable insights to analyze the fine-tuning process.

\paragraph{Flexible Evaluation}

\ifname supports calculating the text similarity scores on the datasets to automatically evaluate models or performing human evaluation by chatting with them.

\paragraph{Multilingual Support}

\ifname provides localization files, facilitating the integration of new languages for rendering the interface. Currently we support three languages: English, Russian and Chinese, which allows a broader range of users to utilize \ifname for fine-tuning LLMs.

\section{Empirical Study}

We systematically evaluate \fwname from two perspectives: 1) the training efficiency in terms of memory usage, throughput and perplexity. 2) the effectiveness of adaptation to downstream tasks.

\subsection{Training Efficiency}

\paragraph{Experimental Setup}

We utilize the PubMed dataset \cite{canese2013pubmed}, which comprises over 36 million records of biomedical literature. We extract around 400K tokens from the abstract of the literature to construct the training corpus. Then we fine-tune the Gemma-2B \citep{team2024gemma}, Llama2-7B and Llama2-13B \citep{touvron2023llama2} models using the generative pre-training objective with various efficient fine-tuning methods. We compare the results of full-tuning, freeze-tuning, GaLore, LoRA and 4-bit QLoRA. After fine-tuning, we calculate the perplexity on the training corpus to evaluate the efficiency of different methods. We also incorporate the perplexities of the pre-trained models as baselines.

In this experiment, we adopt a learning rate of $10^{-5}$, a token batch size of $512$. We fine-tune these models using the 8-bit AdamW optimizer \citep{dettmers2022adamw8bit} in bfloat16 precision with activation checkpointing to reduce the memory footprint. In freeze-tuning, we only fine-tune the last $3$ decoder layers of the model. For GaLore, we set the rank and scale to $128$ and $2.0$, respectively. For LoRA and QLoRA, we attach adapters to all linear layers and set the rank and alpha to $128$ and $256$, respectively. All the experiments are conducted on a single NVIDIA A100 40GB GPU. We enable flash attention in all experiments and Unsloth for LoRA and QLoRA experiments.

\paragraph{Results}

The results about the training efficiency are presented in Table~\ref{tab:efficiency}, where memory refers to the peak memory consumed during training, throughput is calculated as the number of tokens trained per second, and PPL represents the perplexity of the model on the training corpus. Since full-tuning Llama2-13B lead to a memory overflow, the results are not recorded. We observe that QLoRA consistently has the lowest memory footprint because the pre-trained weights are represented in lower precision. LoRA exhibits higher throughput leveraging the optimization in LoRA layers by Unsloth. GaLore achieves lower PPL on large models while LoRA advantages on smaller ones.

\subsection{Fine-Tuning on Downstream Tasks}

\begin{table*}[t]
\centering
\resizebox{1.99\columnwidth}{!}{%
\begin{tabular}{l|ccccc|ccccc|ccccc}
\toprule

&
\multicolumn{5}{|c|}{CNN / DM} &
\multicolumn{5}{|c|}{XSum}     &
\multicolumn{5}{|c}{AdGen}     \\

Model &
Baseline & FT & GaLore & LoRA & QLoRA &
Baseline & FT & GaLore & LoRA & QLoRA &
Baseline & FT & GaLore & LoRA & QLoRA \\

\midrule

ChatGLM3-6B &
18.51 & 22.00 & \underline{22.16} & 21.68 & 21.70 &
16.14 & 26.25 & 26.34 & 26.50 & \underline{26.78} &
14.53 & 19.91 & \underline{20.57} & 20.47 & 20.49 \\

Yi-6B &
16.85 & 22.40 & 22.68 & \underline{22.98} & 22.97 &
18.24 & 27.09 & 28.25 & 28.71 & \underline{29.21} &
13.34 & 19.68 & 20.06 & \underline{20.97} & 20.31 \\

Llama2-7B &
12.94 & \underline{22.87} & 22.40 & 22.70 & 22.61 &
13.89 & 27.69 & 27.64 & \underline{28.80} & 28.05 &
0.61  & \underline{20.51} & 19.61 & 20.29 & 20.45 \\

Mistral-7B &
14.39 & 22.03 & 22.99 & \underline{23.47} & 23.28 &
15.87 & 23.57 & 28.00 & 30.41 & \underline{30.44} &
7.82  & 20.14 & 20.90 & \underline{20.99} & 20.56 \\

Gemma-7B &
15.97 & 22.07 & / & 22.41 & \underline{22.44} & 
15.31 & 25.13 & / & 28.67 & \underline{29.02} &
11.57 & 19.99 & / & \underline{20.62} & 19.81 \\

Qwen1.5-7B &
15.40 & 22.46 & 21.76 & \underline{22.71} & 22.52 &
19.27 & 26.68 & 26.64 & \underline{27.77} & 27.60 &
14.49 & 20.42 & 21.08 & 21.31 & \underline{21.34} \\

Qwen2-7B &
16.46 & 23.20 & / & 23.29 & \underline{23.66} &
19.76 & 26.94 & / & 28.92 & \underline{28.94} &
12.89 & 19.83 & / & \underline{20.96} & 20.86 \\

Llama3-8B &
15.19 & 23.36 & 23.57 & 23.48 & \underline{\bf 24.12} &
17.83 & 26.21 & 30.45 & 30.63 & \underline{\bf 30.94} &
0.22 & 20.28 & 21.27 & \underline{\bf 21.44} & 21.20 \\

\bottomrule
\end{tabular}%
}
\caption{Comparison of the performance (in terms of ROUGE) on specific tasks using different fine-tuning methods in \fwname. The best result of each model is \underline{underlined}, and the best result of each task is in {\bf bold}.}
\label{tab:eval}
\end{table*}

\paragraph{Experimental Setup}

To evaluate the effectiveness of different efficient fine-tuning methods, we compare the performance of various models after fine-tuning on downstream tasks. We construct non-overlapping training set and test set using 2,000 examples and 1,000 examples from three representative text generation tasks, including CNN/DM \citep{nallapati2016cnndm}, XSum \citep{narayan2018xsum} and AdGen \citep{shao2019adgen}, respectively. We select several instruction-tuned models and fine-tune them following the sequence-to-sequence task using different fine-tuning methods. Then we compare the results of full-tuning (FT), GaLore, LoRA and 4-bit QLoRA. After fine-tuning, we calculate the ROUGE score \citep{lin2004rouge} on the test set of each task. We also incorporate the scores of the original instruction-tuned models as baselines.

In this experiment, we set learning rate to $10^{-5}$, batch size to $4$ and maximum input length to $2048$. We fine-tune these models using the 8-bit AdamW optimizer \citep{dettmers2022adamw8bit} in bfloat16 precision with activation checkpointing. For GaLore, we set the rank and scale to $128$ and $2.0$, respectively. For LoRA and QLoRA, we attach adapters to all linear layers and set the rank and alpha to $128$ and $256$, respectively. All the experiments are conducted on NVIDIA A100 40GB GPUs.

\paragraph{Results}

The evaluation results on downstream tasks are shown in Table~\ref{tab:eval}. We report the averaged scores over ROUGE-1, ROUGE-2 and ROUGE-L. Some results of the Gemma-7B and Qwen2-7B \citep{bai2023qwen} models are not included in the table because the GaLore method may not be applicable to them. An interesting finding from the results is that LoRA and QLoRA achieve the best performance in most cases, except for the ChatGLM3-6B \citep{zeng2024chatglm} and Llama2-7B models on the CNN/DM and AdGen datasets. This phenomenon highlights the effectiveness of these efficient fine-tuning methods in adapting LLMs to specific tasks. Additionally, we observe that Llama3-8B achieves the best performance among these models, while Yi-6B \citep{young2024yi} and Mistral-7B \citep{jiang2023mistral} exhibit competitive performance among models of the same size.

\section{Conclusion and Future Work}

In this paper, we demonstrate \fwname, a unified framework for the efficient fine-tuning of LLMs. Through a modular design, we minimize dependencies between the models, datasets and training methods and provide an integrated approach to fine-tune over 100 LLMs with a diverse range of efficient fine-tuning techniques. Additionally, we offer a flexible web UI \ifname, enabling customized fine-tuning and evaluation of LLMs without coding efforts. We empirically validate the efficiency and effectiveness of our framework on language modeling and text generation tasks.

We will consistently keep \fwname synchronous with the state-of-the-art models and efficient fine-tuning techniques. We also welcome contributions from the open-source community. The road map of \fwname including:

(1) Enabling fine-tuning for models that supports a wider range of modalities, \eg, the audio and video modalities \citep{zhu2024languagebind}.

(2) Integrating more parallel training strategies, \eg, sequence parallelism \citep{jacobs2023ulysses} and tensor parallelism \citep{shoeybi2019megatron}.

(3) Exploring stronger fine-tuning methods for conversational models, \eg, self-play \citep{chen2024spin,yuan2024selfreward}.

\section{Broader Impact and Responsible Use}

\fwname has attracted a large number of individuals interested in LLMs to explore the possibility of customizing models. This contributes significantly to the growth of the open-source communities. It is gaining increasing attention and is being featured in Awesome Transformers\footnote{\url{https://github.com/huggingface/transformers/blob/v4.40.0/awesome-transformers.md\#llama-factory}} as a representative of efficient fine-tuning frameworks for LLMs. We anticipate that practitioners build their LLMs upon our framework that bring benefits to society. Adherence to the model license is mandatory when using \fwname for fine-tuning LLMs, thus preventing from any potential misuse.

\section*{Acknowledgements}

This work is supported partly by the National Science and Technology Major Project under Grant 2022ZD0120202, by the National Natural Science Foundation of China (No. U23B2056), by the Fundamental Research Funds for the Central Universities, and by the State Key Laboratory of Complex \& Critical Software Environment.

\bibliography{main}

\begin{table*}[t]
\centering
\resizebox{1.7\columnwidth}{!}{%
\begin{tabular}{llll}
\toprule
Model & Variant & Organization & Release Date \\
\midrule
Llama \citep{touvron2023llama} & 7B/13B/33B/65B & Meta AI & Feb. 2023 \\
Llama 2 \citep{touvron2023llama2} & 7B/13B/70B & Meta AI & Jul. 2023 \\
Llama 3 \citep{meta2024llama3} & 8B/70B & Meta AI & Apr. 2024 \\
Aya 23 \citep{aryabumi2024aya} & 8B/35B & Cohere For AI & May 2024 \\
Baichuan \cite{yang2023baichuan} & 7B/13B & Baichuan Inc & Jun. 2023 \\
Baichuan2 \citep{yang2023baichuan} & 7B/13B & Baichuan Inc & Sep. 2023 \\
BLOOM \citep{le2022bloom} & 560M/3B/7.1B & BigScience & May 2022 \\
BLOOMZ \citep{le2022bloom} & 560M/3B/7.1B & BigScience & Sep. 2022 \\
ChatGLM2 \citep{zeng2024chatglm} & 6B & Zhipu AI & Jun. 2023 \\
ChatGLM3 \citep{zeng2024chatglm} & 6B & Zhipu AI & Oct. 2023 \\
ChineseLLaMA2 \citep{cui2023efficient} & 3B/7B/13B & HFL & Jul. 2023 \\
CodeGemma \citep{team2024codegemma} & 2B/7B & Google & Apr. 2024 \\
DeepSeek-Coder \citep{guo2024deepseekcoder} & 6.7B/7B/33B & DeepSeek AI & Oct. 2023 \\
DeepSeek-Coder-V2 \citep{zhu2024deepseekcoderv2} & 16B/236B & DeepSeek AI & Jun. 2024 \\
DeepSeek-LLM \citep{bi2024deepseek} & 7B/67B & DeepSeek AI & Nov. 2023 \\
DeepSeek-Math \citep{shao2024deepseekmath} & 7B & DeepSeek AI & Feb. 2024 \\
DeepSeek-MoE \citep{dai2024deepseekmoe} & 16B & DeepSeek AI & Jan. 2024 \\
DeepSeek-V2 \citep{liu2024deepseekv2} & 16B/236B & DeepSeek AI & May 2024 \\
Falcon \citep{almazrouei2023falcon} & 7B/11B/40B/180B & TII & Apr. 2023 \\
Gemma \citep{team2024gemma} & 2B/7B & Google & Feb. 2024 \\
Gemma 2 \citep{team2024gemma} & 9B/27B & Google & Jun. 2024 \\
GLM-4 \citep{zeng2024chatglm} & 9B & Zhipu AI & Jun. 2024 \\
InternLM \citep{team2023internlm} & 7B/20B & Shanghai AI Lab & Jul. 2023 \\
InternLM2 \citep{team2023internlm} & 1.8B/7B/20B & Shanghai AI Lab & Jan. 2024 \\
LLaVA \citep{liu2024llava} & 7B/13B & / & Apr. 2023 \\
MiniCPM \citep{hu2024minicpm} & 2B & ModelBest Inc & Jan. 2024 \\
Mistral \citep{jiang2023mistral} & 7B & Mistral AI & Sep. 2023 \\
Mixtral \citep{jiang2023mistral} & 8x7B/8x22B & Mistral AI & Dec. 2023 \\
OLMo \citep{groeneveld2024olmo} & 1B/7B & Allen AI & Jan. 2024 \\
OpenChat \citep{wang2023openchat} & 7B & OpenChat & Jul. 2023 \\
Orion \citep{chen2024orion} & 14B & OrionStar & Jan. 2024 \\
PaliGemma \citep{google2024paligemma} & 3B & Google & May. 2024 \\
Phi-1.5 \citep{li2023textbooks} & 1.3B & Microsoft & Sep. 2023 \\
Phi-2 \citep{li2023textbooks} & 2.7B & Microsoft & Dec. 2023 \\
Phi-3 \citep{abdin2024phi3} & 3.8B/7B/14B & Microsoft & Apr. 2024 \\
Qwen \citep{bai2023qwen} & 1.8B/7B/14B/72B & Alibaba Cloud & Sep. 2023 \\
Qwen1.5 \citep{bai2023qwen} & 1.8B/7B/14B/72B & Alibaba Cloud & Jan. 2024 \\
Qwen2 \citep{bai2023qwen} & 0.5B/1.5B/7B/72B & Alibaba Cloud & Jun. 2024 \\
SOLAR \citep{kim2023solar} & 10.7B & Upstage AI & Dec. 2023 \\
Skywork \citep{wei2023skywork} & 13B & Skywork & Oct. 2023 \\
StarCoder2 \citep{lozhkov2024starcoder} & 3B/7B/15B & BigCode & Feb. 2024 \\
TeleChat \citep{wang2024telechat} & 7B/12B & Telecom & Mar. 2024 \\
Vicuna1.5 \citep{zheng2023fastchat} & 7B/13B & LMSYS & Jul. 2023 \\
Yi \citep{young2024yi} & 6B/9B/34B & 01.AI & Nov. 2023 \\
Yi-1.5 \citep{young2024yi} & 6B/9B/34B & 01.AI & May 2024 \\
Yuan2 \citep{wu2023yuan} & 2B/51B/102B & IEIT & Dec. 2023 \\
Zephyr \citep{tunstall2023zephyr} & 7B & Hugging Face H4 & Oct. 2023 \\
\bottomrule
\end{tabular}%
}
\caption{List of supported models by \fwname. Note that the models from unknown sources are not included in this list.}
\label{tab:models}
\end{table*}

\end{document}